\documentclass[letterpaper, 10 pt, conference]{ieeeconf}
\IEEEoverridecommandlockouts 
\usepackage{graphicx}
\usepackage{caption, subcaption}
\usepackage{amsmath,amssymb,enumerate}

\usepackage{amsthm}
\usepackage{mathtools}
\usepackage{breqn}
\usepackage[usenames,dvipsnames,svgnames,table, xcdraw]{xcolor}
\usepackage[colorlinks=true,pdfpagemode=UseNone,citecolor=black,linkcolor=black,urlcolor=BrickRed]{hyperref}
\usepackage{algorithm, algpseudocode}

\usepackage{blindtext}
\usepackage{gensymb}
\usepackage{xparse}
\usepackage{lipsum}
\usepackage{mathrsfs}
\usepackage[mathscr]{euscript}
\usepackage{times}
\usepackage[noadjust]{cite} 
\usepackage{multicol}
\usepackage{amsfonts}
\usepackage[utf8]{inputenc}
\usepackage[T1]{fontenc}
\usepackage{textcomp}
\usepackage{balance}
\usepackage{soul}
\usepackage{multirow}
\usepackage{booktabs}
\usepackage{sidecap}
\usepackage{makecell}
\usepackage{array,float}

\renewcommand{\thefigure}{\arabic{figure}}

\definecolor{free}{rgb}{0.9608, 0.5882, 0.3922}
\definecolor{building}{rgb}{0.9608, 0.9020, 0.3922}
\definecolor{barrier}{rgb}{0.5882, 0.2353, 0.1176}
\definecolor{other}{rgb}{0.7059, 0.1176, 0.3137}
\definecolor{pedestrian}{rgb}{1, 0.3137, 0.3922}
\definecolor{pole}{rgb}{0.1176, 0.1176, 1}
\definecolor{road}{rgb}{0.7843, 0.1569, 1}
\definecolor{ground}{rgb}{0.3529, 0.1176, 0.5882}
\definecolor{sidewalk}{rgb}{1, 0, 1}
\definecolor{vegetation}{rgb}{1, 0.5882, 1}
\definecolor{vehicles}{rgb}{0.2941, 0, 0.2941}
\definecolor{fence}{rgb}{0.1961, 0.4706, 1}
\definecolor{sign}{rgb}{0, 0, 1}

\definecolor{scarColor}{rgb}{0.9608, 0.5882, 0.3922}
\definecolor{sbicycleColor}{rgb}{0.2608, 0.9020, 0.3922}
\definecolor{smotorcycleColor}{rgb}{0.5882, 0.2353, 0.1176}
\definecolor{struckColor}{rgb}{0.7059, 0.1176, 0.3137}
\definecolor{sothervehicleColor}{rgb}{1, 0.3137, 0.3922}
\definecolor{spersonColor}{rgb}{0.1176, 0.1176, 1}
\definecolor{sbicyclistColor}{rgb}{0.7843, 0.1569, 1}
\definecolor{smotorcyclistColor}{rgb}{0.3529, 0.1176, 0.5882}
\definecolor{sroadColor}{rgb}{0.5, 0, 1}
\definecolor{sparkingColor}{rgb}{1, 0.5882, 1}
\definecolor{ssidewalkColor}{rgb}{0.2941, 0, 0.2941}
\definecolor{sothergroundColor}{rgb}{0.2941, 0, 0.6863}
\definecolor{sbuildingColor}{rgb}{0, 0.7843, 1}
\definecolor{sfenceColor}{rgb}{0.1961, 0.4706, 1}
\definecolor{svegetationColor}{rgb}{0, 0.6863 ,0}
\definecolor{strunkColor}{rgb}{0, 0.2353, 0.5294}
\definecolor{sterrainColor}{rgb}{0.3137, 0.9412, 0.5882}
\definecolor{spoleColor}{rgb}{0.5882, 0.7412, 1}
\definecolor{strafficsignColor}{rgb}{0, 0, 1}

\graphicspath{ {./media/} }
\usepackage{svg}
\usepackage{multirow}
\usepackage{amssymb}
\usepackage{pifont}
\usepackage[flushleft]{threeparttable}
\usepackage{subcaption}
\usepackage{wrapfig}

\renewcommand{\baselinestretch}{0.99}

\begin{document}

\title{LiRaFusion: Deep Adaptive LiDAR-Radar Fusion for 3D Object Detection}

\author{Jingyu Song, Lingjun Zhao, and Katherine A. Skinner
\thanks{This work is supported by the Ford Motor Company via the Ford-UM
Alliance under award N028603.}
\thanks{J. Song, L. Zhao, and K. Skinner are with the Department of Robotics, University of Michigan, Ann Arbor, MI, USA {\tt\footnotesize jingyuso,lingjunz,kskin@umich.edu}.}
}

\date{April 2023}

\maketitle

\begin{abstract}
   We propose LiRaFusion to tackle LiDAR-radar fusion for 3D object detection to fill the performance gap of existing LiDAR-radar detectors. To improve the feature extraction capabilities from these two modalities, we design an early fusion module for joint voxel feature encoding, and a middle fusion module to adaptively fuse feature maps via a gated network. We perform extensive evaluation on nuScenes to demonstrate that LiRaFusion leverages the complementary information of LiDAR and radar effectively and achieves notable improvement over existing methods.
\end{abstract}

\section{Introduction}
\label{sec:intro}
Autonomous vehicles (AVs) are expected to accurately perceive the surrounding environment to enable effective and safe planning and control across a variety of scenarios and environmental conditions~\cite{3d_object_detection_review_new_outlook, survey_detection_multi, wilson2022motionsc, wilson2023convolutional, xu2023online}.
An important part of the perception task is to precisely localize the objects in the surrounding environment. A common representation for these objects is a set of 3D bounding boxes that have locations, sizes and classes \cite{kitti, nuscenes}.
Despite various combinations of sensor configurations on AVs, many object detection algorithms rely on LiDAR and cameras due to their dense returns~\cite{survey_detection_multi, survey_object_detection_2022,3d_object_detection_review_new_outlook,zhang2023coda, second, pointpainting}.

Still, LiDAR systems and cameras are sensitive to varying weather and lighting conditions, so AVs can suffer significantly from downgraded perceptual capability in these scenarios. To tackle this problem, recent research has focused on leveraging radar systems, which have automotive-grade design that ensures robust performance under various conditions~\cite{survey_deep_radar, radarnet, radar_voxel_fusion, ezfusion}. Additional benefits of radars include their low cost, long detection range and Doppler effect information (i.e., velocity of captured targets). 
Therefore, it is of great significance to design a model that could effectively leverage radar for 3D object detection~\cite{survey_detection_multi, radar_voxel_fusion, radarnet}. 

Existing detectors with radars can be categorized into single-modality methods \cite{radar_detection_improved} and fusion-based methods \cite{radarnet,ezfusion}.
Recent works \cite{mvdnet, mod_ago_learning_lidar_radar} have achieved impressive detection accuracy when fusing LiDAR and radars on the Oxford Radar RobotCar dataset \cite{oxford_radar_dataset}, which has high-resolution radar data. However, this dataset uses a spinning radar, which lacks Doppler information and has increased cost \cite{survey_deep_radar}. Among the popular datasets for 3D object detection \cite{mmdet3d2020, kitti, survey_object_detection_2022, survey_deep_radar}, nuScenes \cite{nuscenes} stands out because it is large-scale and has a complete sensor suite including radars. nuScenes represents radar data as object lists, which is a common representation that could also be interpreted as a very sparse point cloud with additional feature attributes from radar on-board signal processing \cite{survey_deep_radar, radarnet}. The common challenges with this dataset are the sparsity and noise of the radar data. Consequently, single-modality radar detectors fail to achieve reliable performance. Some fusion-based detectors \cite{FUTR3D, deepfusion_lcr} suffer from downgraded performance when adding radar to LiDAR-only or LiDAR-camera fusion detectors, while some detectors \cite{ezfusion, radarnet, sparse_pointnet} have to enforce hard constraints such as limiting detection to specific classes or limiting detection range to achieve improvement in performance. Our work seeks to fill the gap in the current literature by improving the design of fusion architecture for LiDAR and radar by leveraging their shared point cloud representation. 

\begin{figure}[t!]
\begin{center}

  \includegraphics[width=1.0\linewidth]{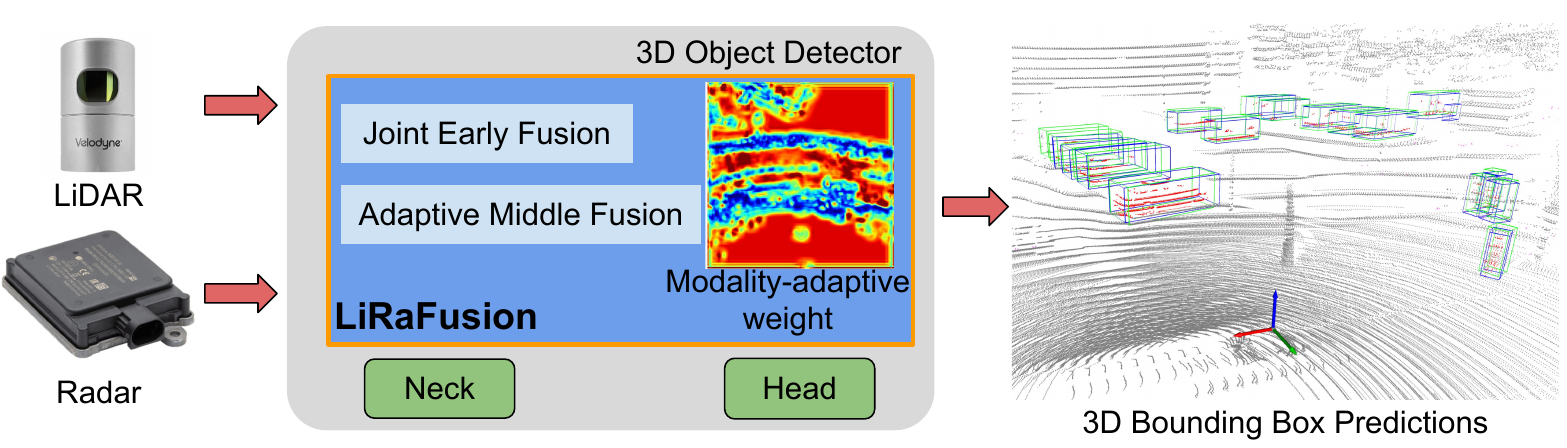}
  \caption{We propose LiRaFusion to efficiently leverage the complementary information of LiDAR and radar for 3D object detection.}
    \label{fig:intro_lira}
\end{center}
\vspace{-8mm}
\end{figure}
In this work, we propose a novel LiDAR-radar fusion detector, LiRaFusion (Fig.~\ref{fig:intro_lira}). Our main contributions are: (i) a novel joint feature extractor for effective LiDAR-radar fusion, (ii) the first introduction of the adaptive gated network into LiDAR-radar fusion for object detection with novel improvement considering the bird’s-eye-view (BEV) feature space, and (iii) extensive evaluation on open source datasets and detectors that demonstrate improvement over existing LiDAR-radar fusion methods.
As most existing detectors follow the backbone-neck-head design \cite{second, centerpoint, FUTR3D}, LiRaFusion can be directly integrated into existing methods by serving as the backbone to enable more modality configurations, which is validated by extending it to LiDAR-camera-radar fusion. Code will be made available on the project website.\footnote{\url{https://github.com/Song-Jingyu/LiRaFusion}} 

\section{Related Work}

\subsection{Radar Datasets for Autonomous Driving}
\label{sec:radar_dataset_related_works}
Data is the key component for enabling development of learning-based object detectors. However, radar data was rarely available in early public autonomous driving datasets~\cite{kitti, waymo_dataset, mmdet3d2020}. Recently, more datasets with radar data have become accessible to researchers. In these datasets, radar data usually has two representations. The most common representation is the point cloud, in which each point represents an object in the object list output by many off-the-shelf radars with on-board processing algorithms such as Constant False Alarm Rate (CFAR)~\cite{survey_deep_radar, radarnet}. This representation is available on datasets such as nuScenes~\cite{nuscenes}, aiMotive~\cite{aimotive}, and Zendar~\cite{zendar}. There are also datasets that directly use the raw data from radars \cite{oxford_radar_dataset, sheeny2021radiate}. The raw data has denser information but the lack of CFAR processing leads to increased noise. The spinning radar configuration in \cite{oxford_radar_dataset} loses Doppler information (e.g., velocities of the captured targets), which is important to understand the scene. Another challenge brought by this representation is the difficulty of annotating the data \cite{survey_deep_radar}. In this work we leverage the nuScenes dataset~\cite{nuscenes} because of its full coverage on sensors and driving scenarios, accurate annotations, and popularity, which allows us to compare this work to many existing works. The proposed fusion architecture can be transferred to other datasets that have similar point cloud representation.

\subsection{Multi-modality 3D Object Detection}
3D object detection is an important part of the perception system for AVs \cite{3d_object_detection_review_new_outlook}. The main objective is to assign a class label and a 3D bounding box for each detected object in the scene~\cite{survey_object_detection_2022}. Common sensors used for 3D object detection include LiDAR and cameras. 
Though several single-modality object detectors with LiDAR \cite{centerpoint} or cameras \cite{li2022bevformer} achieve impressive results on KITTI \cite{kitti} or nuScenes \cite{nuscenes} benchmarks, multi-modality object detectors have recently shown promise in leveraging complementary information to improve robustness and accuracy~\cite{survey_detection_multi,centerpoint, mvx_net, pointfusion, deepfusion_lcr,clocs,pointpainting}. 
LiDAR-camera fusion is the most common configuration. However, these two sensors have shared drawbacks (e.g., sparse information at long range, lacking velocity estimation) that could be compensated by radars that are commonly deployed on AVs \cite{survey_deep_radar, radarnet}. Radar-only detectors usually fail to overcome the data sparsity to perform comparably to camera- or LiDAR-based detectors \cite{rodnet, radar_detection_improved}. 

One recent trend is to fuse radar with one or several other sensors. Existing fusion configurations include camera-radar (CR), LiDAR-radar (LR) and LiDAR-camera-radar (LCR)~\cite{centerfusion, john2019rvnet, radarnet, radar_voxel_fusion, seeing_through_fogs, FUTR3D, ezfusion}. As the main focus of this work is fusing LiDAR and radar in the shared point cloud representation, we mainly compare our methods with FUTR3D \cite{FUTR3D} and EZFusion \cite{ezfusion} because they are the most recent state-of-the-art detectors supporting LR fusion on the  nuScenes \cite{nuscenes} dataset. In FUTR3D \cite{FUTR3D}, though LR and LCR configurations are supported, they suffer from downgraded performance when compared with LiDAR-only (LO) or LC fusion. We argue the failure could come from the simple MLP-style feature extractor for radars and lack of joint feature fusion before sampling features for query points. Our proposed method aims to address these limitations with the proposed adaptive fusion framework.
EZFusion \cite{ezfusion} is built on CenterPoint \cite{centerpoint} by adding radar feature projection for the LiDAR points.
In EZFusion \cite{ezfusion}, though its LR fusion shows improvement over its LO configuration (equivalent to CenterPoint \cite{centerpoint}), its partial moving class setting, which uses only the $7$ moving classes out of 10 classes in the nuScenes benchmark, has more limited application for practical deployment since the missed static classes are also vital for keeping AVs safe. Our proposed method achieves further improvement over EZFusion under the same partial classes setting and is demonstrated on the complete class setting, which has stronger potential in application since it can account for both static and moving object classes.

\subsection{Gated Network for Sensor Fusion} 
Fusing information from different modalities requires sophisticated architecture design and there are multiple prior works that addressed this challenge~\cite{deep_multi_detection_segmentation_survey}. Among them, projection, addition and concatenation are common practices \cite{pointpainting, deepfusion_lcr, ezfusion, radar_voxel_fusion}. Though these methods demonstrate improvement on multi-modality fusion, they are not learning-based and lack adaptivity. 
To account for this issue, researchers have turned to gated networks when different feature maps are fused. This process is also named as the mixture of experts because the backbones used for different modalities are considered different expert networks. This method is first introduced in \cite{jacobs1991adaptive}, in which the expert network is defined as a domain-specific neural network to process a single sensing modality, and the gating network is a weighting neural network that selects useful features among the outputs of the expert networks. This idea has been leveraged by the perception community as more works have focused on using different modalities~\cite{survey_detection_multi, 3dcvf, adaptive_gated_depth_estimation, adaptive_cooperative_perception}. For instance, in 3D object detection, 3D-CVF \cite{3dcvf} proposes an LC fusion network using a gated network. Extensive evaluation is conducted on the KITTI \cite{kitti} and nuScenes \cite{nuscenes} datasets, which demonstrates the effectiveness of the gated network. 

 The success of fusing LiDAR and camera modalities through the gated network motivates us to adapt this method for the LR sensor configuration. The gated network is able to learn adaptive weights for different expert networks so the model can learn to be robust to noise from individual experts. We find this feature is of great significance in the LR fusion since the radars are more noisy, which can degrade the performance of multi-modal detectors if the information is not properly handled. In our proposed work, we extend the original gated network design~\cite{GIF,3dcvf} by making it channel-wise so that each channel of the BEV map has an adaptive weight. According to our best knowledge, LiRaFusion is the first to introduce the gated network into LR perception.

\section{Method}
\begin{figure*}[h!]
    \includegraphics[width=0.95\linewidth]{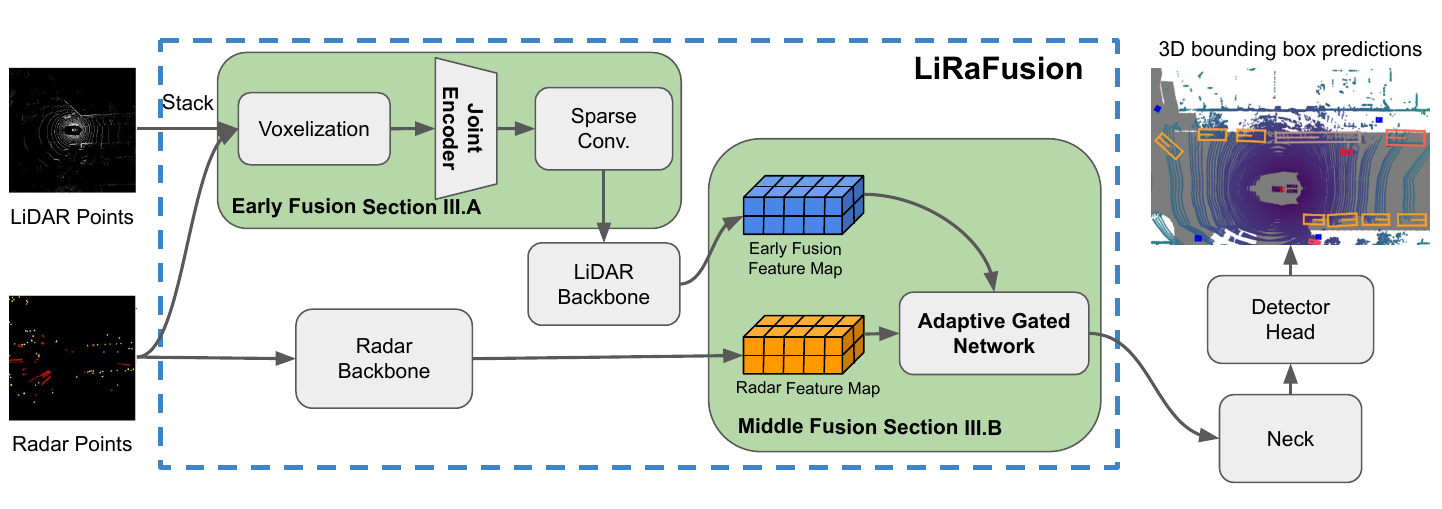}
    \centering
    \caption{Overview of the architecture of LiRaFusion. Our main contributions, shown as bold text, mainly include a joint voxel feature encoder to extract per-voxel features from the stacked point cloud, and a gated network to learn weights for each input feature map to fuse them adaptively.
    }
    \label{fig:overview}
    \vspace{-6mm}
\end{figure*} 
\label{Sec: Method}
The goal of our method, LiRaFusion, is to achieve more effective feature extraction and fusion for LiDAR and radar data for 3D object detection (Fig.~\ref{fig:overview}).
The inputs to LiRaFusion are a LiDAR point cloud and a radar point cloud. 
One stream stacks these two point clouds as the input to the proposed early fusion module. The early fusion module processes the denser point cloud with the proposed joint feature encoder and a common sparse 3D convolution encoder. Its output is then fed into a common LiDAR backbone to obtain the feature map. In this work, we use the VoxelNet following \cite{FUTR3D, centerpoint}. The other stream uses the PointPillars \cite{pointpillars} backbone to process the radar points, taking advantage of the pillars since the height measurement for radar points are noisy \cite{radarnet, survey_deep_radar}. The output is a radar feature map. The output feature maps from these two streams can be considered as two experts, which are further fused with the proposed gated network in the middle fusion module. The middle fusion module learns the adaptive weights for these two feature maps and then concatenates the weighted feature maps together. The concatenated feature map is passed into the Feature Pyramid Network (FPN) \cite{feature_pyramid_network_FPN} neck and the detector head to generate predictions. Our main contributions are the novel architectures for the early fusion and middle fusion modules.  LiRaFusion is an enhanced backbone for LR feature extraction so it can be extended to LCR configuration as well. Technical details of each module are discussed in the following subsections.

\begin{figure}[h!]
\begin{center}
  \includegraphics[width=1.0\linewidth]{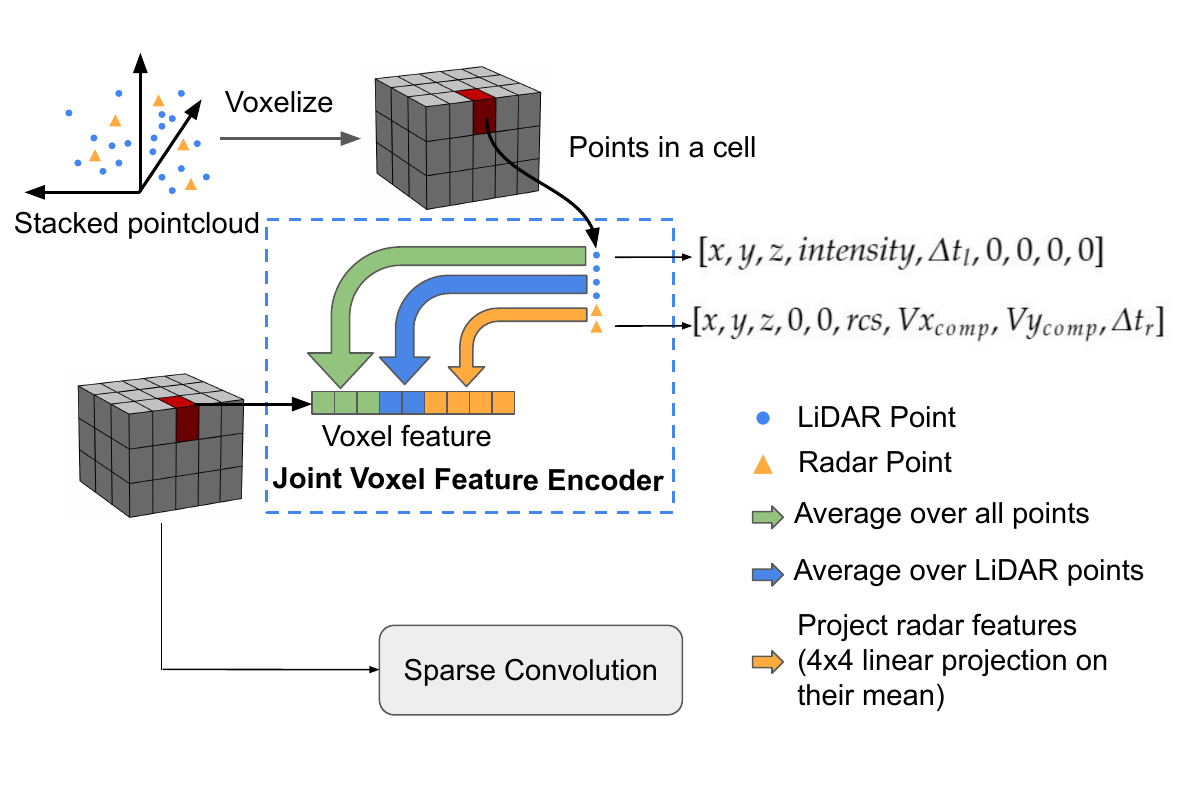}
  \caption{The network architecture early fusion module. We stack the loaded LiDAR and radar points by zero-padding them to the same number of dimensions before feeding into the proposed joint voxel feature encoder. 
   }
  \label{fig:early}
\end{center}
\vspace{-6mm}
\end{figure}
\subsection{Early Fusion}
\label{sec:early_fusion_method}
We design an early fusion block to fuse the LiDAR and radar points at an early stage to extract features for each voxel cell, as shown in Fig.~\ref{fig:early}. Unlike \cite{radarnet, aimotive, sparse_pointnet}, which ignore features such as LiDAR intensity, Radar Cross-section (RCS), and velocity, we keep these features since LiDAR intensity and RCS are helpful to classify objects, and velocity information is important to distinguish static or dynamic objects and predict the velocity and rotation. For the LiDAR input, we keep the point intensity and the captured time offset ($\Delta t_l$) from the current frame as we accumulate multiple sweeps. For the radar points, we keep the RCS, compensated velocities ($V_{x_{\text{comp}}}, V_{y_{\text{comp}}}$) and the time offset ($\Delta t_r$). We use zero-padding to match the dimension of these points to merge them. After voxelization, we use the proposed joint voxel feature encoder to extract features for each voxel cell. We follow the simplified voxel feature encoder of VoxelNet \cite{second} in MMDetection3D \cite{mmdet3d2020} to set the number of the voxel feature dimensions to be the same as the input points. The first $3$ feature dimensions represent the location of the centroid of this cell, which is computed by taking the mean locations of all the points in this voxel cell. The following $2$ feature dimensions correspond to features from LiDAR so we average over all LiDAR points. The last $4$ feature dimensions correspond to the radar features. We pass the mean of the radar features to a $4\times4$ linear layer to enable to network to learn an appropriate way to handle the radar features. We only process non-empty voxel cells. For cells that do not have radar points, which is common due to sparsity, we leave the last $4$ dimensions of these cells as zero. After obtaining the voxel features, we apply standard sparse convolution and further process its output with a standard LiDAR backbone, VoxelNet \cite{second}. To simplify the terminology, we refer to the output of the early fusion stream as the LiDAR feature map in the following sections. Though radar data has already been fused when encoding the voxel feature, due to the sparsity of radar data, most information in the encoded voxel features is from LiDAR. We further fuse it with the radar feature map at a higher level with the proposed middle fusion module.

\begin{figure}[h!]
\begin{center}
  \includegraphics[width=1.0\linewidth]{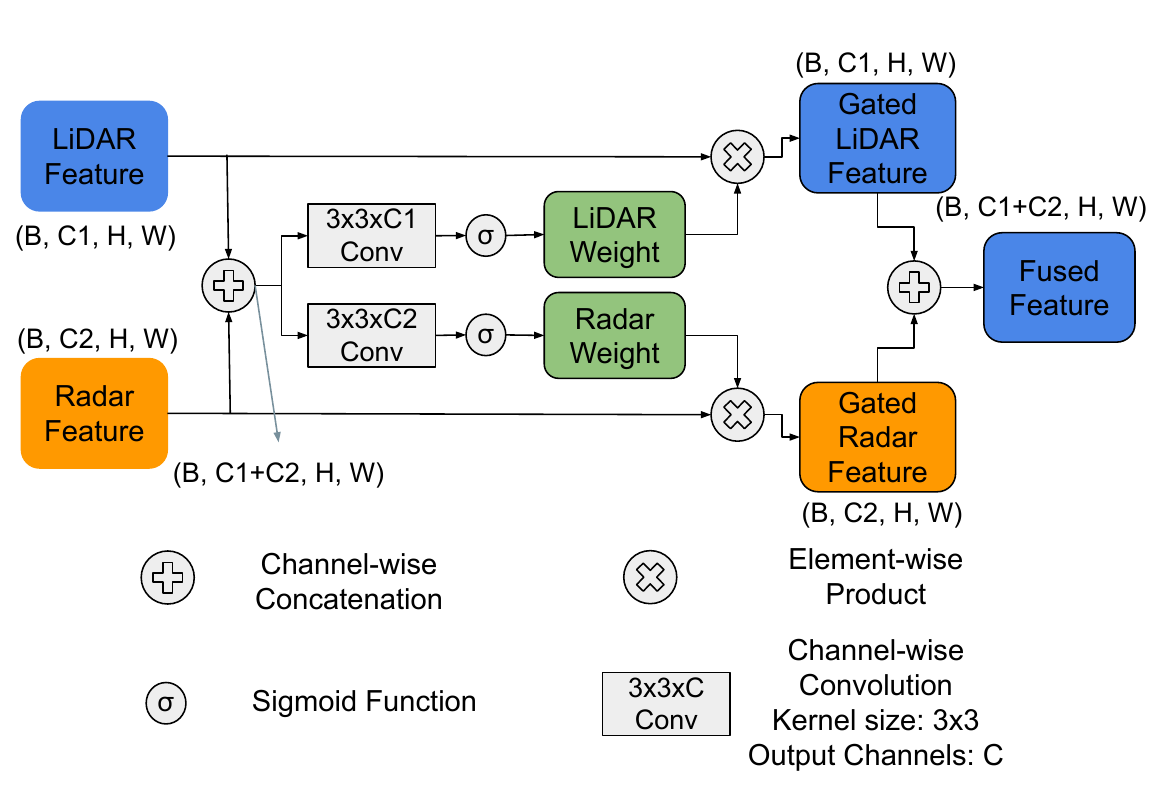}
  \caption{The network architecture for the middle fusion module. In this module, by applying a channel-wise convolution and a sigmoid function to the concatenated LiDAR-radar feature map, the network generates adaptive weights for LiDAR and radar separately. Then the input LiDAR and radar feature maps are element-wise multiplied with the weights before being concatenated as a fused LiDAR-radar feature map.}
  \label{fig:middle}
\end{center}
\vspace{-6mm}
\end{figure}

\subsection{Middle Fusion}
In order to perform adaptive sensor fusion on the feature maps from different modalities, we refer to \cite{GIF, 3dcvf} for designing the gated network. To the best of our knowledge, we are the first to bring the adaptive gated network to the field of LR fusion for 3D object detection. We improve the existing gated network by enabling it to adaptively learn the weight over the channel dimension. Specifically, the generated adaptive weight for the input feature maps are currently in the shape $B \times C \times H \times W$ instead of the previous channel-constant style in the shape $B \times 1 \times H \times W$. This change enables the gated network to weight and extract the LiDAR and radar features in a more flexible way. Intuitively, the feature maps are all in bird's-eye-view resulting from being flattened over the $z$ axis. By proposing a channel-specific weight, we improve the capability of the network to exploit the spatial knowledge from the input experts.

The design of the adaptive gated network is shown in Fig.~\ref{fig:middle}. The input LiDAR feature map $(B \times C_{1} \times H \times W)$ and input radar feature map $(B \times C_{2} \times H \times W)$ are first concatenated in the channel dimension. Then the concatenated feature map is passed to a convolution block that is followed by a sigmoid function. Notably, the output feature dimension of the convolution blocks for LiDAR and radar modalities is set to match the dimension of the input feature maps ($C_{1}$ for LiDAR and $C_{2}$ for radar). The learned adaptive weights are applied to the original input feature maps through an element-wise product operation. The obtained gated LiDAR and radar features are further concatenated together along the feature dimension as a fused feature map. The resulting fused feature map has the shape $B \times (C_{1}+C_{2}) \times H \times W$ and serves as the output of the middle fusion block.

\section{Experiments}

\subsection{Experiment Design}
As mentioned in Section \ref{sec:radar_dataset_related_works}, we evaluate our method on the nuScenes dataset \cite{nuscenes}. We follow the official split that has 700 scenes for training and 150 scenes for validation. It only has annotations for the key-frames (samples), but also provides non-key-frames (sweeps) without annotations. We follow common practices in \cite{mmdet3d2020, FUTR3D, ezfusion} to load multiple sweeps into a current sample frame to increase the data density, while appending a time difference channel for data from each sweep as additional temporal information.

We use the mean-Average-Precision (mAP) and nuScenes Detection Score (NDS)~\cite{nuscenes} as the main evaluation metrics. We compare LiRaFusion with the existing LR and LCR detectors. In addition, we group predictions to break down the improvement of fusing radars. We also report runtime and additional TP (True Positive) metrics from nuScenes~\cite{nuscenes}. Ablation studies are conducted to validate our model design.

As mentioned previously, many existing detectors \cite{centerpoint, second, FUTR3D, mmdet3d2020} follow the backbone-neck-head design. Therefore, LiRaFusion can be integrated by replacing their backbones with LiRaFusion and keeping the same neck and head. As the baselines we choose -- FUTR3D \cite{FUTR3D} and EZFusion \cite{ezfusion} -- are initially proposed to work with different heads, we group our experiments based on the detector heads. Inspired by \cite{detr, detr3d}, FUTR3D \cite{FUTR3D} uses a transformer-based head, which is referred to as TransHead. EZFusion \cite{ezfusion} uses the same head as CenterPoint \cite{centerpoint}, which is referred to as CenterHead. We choose the LO and LR configurations for FUTR3D as the baselines, denoted as FUTR3D-LO and FUTR3D-LR. We implement the LR fusion strategy in EZFusion with both heads, which is named as EZFusion-LR$^{*}$, where $^*$ represents our re-implementation. The other group of experiments with CenterHead~\cite{centerpoint} focuses on $7$ moving classes out of $10$ complete classes in the nuScenes official benchmark for consistency with results reported in EZFusion \cite{ezfusion}. We re-train the original CenterPoint \cite{centerpoint} with $7$ classes and name it as CenterPoint-7. Since FUTR3D-LO and CenterPoint are the state-of-the-art LO detectors, we include them in our comparison to demonstrate the improvement obtained with fusing radar data. We implemented LiRaFusion based on the MMDetection3D framework \cite{mmdet3d2020}. More implementation and training details can be found on the project website.

\subsection{Results and Comparison}

\begin{table*}[h!]
    \caption{Results with TransHead~\cite{FUTR3D} evaluated on nuScenes \emph{val} set. EZFusion-LR$^*$ represents our re-implementation of \cite{ezfusion}. All values are percentages. No model ensemble or test-time augmentation is used.}
    \begin{threeparttable}
\begin{tabular}{c|c|c|ccccccccccc}
\hline
\multirow{2}{*}{Method} & \multirow{2}{*}{Sensor} & \multirow{2}{*}{NDS $\uparrow$} & \multicolumn{11}{c}{AP (Average Precision) $\uparrow$}                                                                                                                         \\
                        &                         &                                 & mean (mAP)     & Car           & Ped           & Bicycle       & Bus           & Barrier       & TC            & Truck         & Trailer       & Moto          & CV            \\ \hline
FUTR3D-LO               & LO                      & 65.74                           & 59.39          & 84.3          & 81.4          & 49.0          & 65.4          & \textbf{62.4} & 64.2          & 53.5          & \textbf{41.8} & 66.4          & 25.5          \\
FUTR3D-LR               & LR                      & 65.37                           & 58.08          & 83.8          & 81.2          & \textbf{49.8} & 65.4          & 60.4          & 60.6          & 51.0          & 41.0          & 65.1          & 22.6          \\
EZFusion-LR$^*$         & LR                      & 65.77                           & 59.24          & 84.6          & 81.7          & 47.3          & 69.1          & 62.0          & 65.7          & 52.2          & 39.7          & 66.9          & 23.3          \\
LiRaFusion (ours)         & LR                      & \textbf{66.69}                  & \textbf{60.11} & \textbf{85.6} & \textbf{82.2} & 46.9          & \textbf{69.6} & 61.2          & \textbf{66.0} & \textbf{54.0} & 40.7          & \textbf{68.1} & \textbf{26.7}
\end{tabular}
\begin{tablenotes}
\footnotesize
\item Abbreviations: pedestrian (Ped), traffic cone (TC), motorcycle (Moto), and construction vehicle (CV).

\end{tablenotes}
\end{threeparttable}
\vspace{-4mm}
\label{tab:futr3d_results}
\end{table*}

\begin{table*}[t!]
\centering
\caption{Results with CenterHead~\cite{centerpoint} evaluated on nuScenes \emph{val} set. We train these networks with $7$ moving classes. EZFusion-LR stands for the results in its original paper \cite{ezfusion}. $^*$ represents our re-implementation. In this experiment group, test-time augmentation was used to keep consistency with EZFusion \cite{ezfusion}. No model ensemble is used.}
\begin{threeparttable}
\begin{tabular}{c|c|c|cccccccc}
\hline
\multirow{2}{*}{Method} & \multirow{2}{*}{Sensor} & \multirow{2}{*}{NDS $\uparrow$} & \multicolumn{8}{c}{AP (Average Precision) $\uparrow$}                                                                          \\
                        &                         &                                 & mean (mAP)     & Car           & Ped           & Bicycle       & Bus           & Truck         & Trailer       & Moto          \\ \hline
CenterPoint-7           & LO                      & 69.41                           & 61.38          & 86.0          & 79.1          & 40.3          & 68.7          & \textbf{57.7} & 37.1          & 60.7          \\
EZFusion-LR             & LR                      & N/A                             & 63.21          & N/A           & N/A           & N/A           & N/A           & N/A           & N/A           & N/A           \\
EZFusion-LR$^*$         & LR                      & 69.85                           & 61.85          & 86.0          & 79.1          & 43.6          & 70.3          & 57.4          & 35.6          & 60.8          \\
LiRaFusion (ours)         & LR                      & \textbf{72.16}                  & \textbf{65.18} & \textbf{86.8} & \textbf{79.4} & \textbf{54.8} & \textbf{70.4} & 57.5          & \textbf{38.7} & \textbf{68.7}
\end{tabular}
\footnotesize
\end{threeparttable}
\label{tab:centerpoint_results}
\vspace{-5mm}
\end{table*}

We perform a comparison of our method with several state-of-the-art LR fusion networks with CenterHead~\cite{centerpoint} and TransHead~\cite{FUTR3D} separately on the nuScenes dataset~\cite{nuscenes}. Table \ref{tab:futr3d_results} shows the results of all models trained with TransHead~\cite{FUTR3D}. We can see that FUTR3D-LR performs worse than FUTR3D-LO, which proves that ineffective design of LR fusion strategy could actually harm performance. The re-implemented EZFusion with TransHead (EZFusion-LR$^*$) fails to achieve further improvement over FUTR3D-LO. Our model, LiRaFusion, achieves the best results on the nuScenes validation set in terms of NDS and mAP. It also shows impressive improvement on certain classes such as car and pedestrian, the top two most frequent classes, which are critical for AVs to detect in the scene in order to operate effectively and safely.

Table \ref{tab:centerpoint_results} shows the results for models trained with CenterHead using the $7$ moving classes for consistency with EZFusion \cite{ezfusion}. Similarly to EZFusion, we re-train CenterPoint with the 7-class setting as a baseline LO detector and name it as CenterPoint-7. Though EZFusion-LR$^*$ achieves considerable improvement over CenterPoint-7, there is a small gap between EZFusion-LR$^*$ and the reported results in EZFusion \cite{ezfusion} so we include both of them in the table. LiRaFusion is the top performer in terms of NDS and mAP. Similar to its performance with TransHead~\cite{FUTR3D}, LiRaFusion achieves impressive improvement in terms of AP over almost all classes. 

\begin{table}[t!]
    \centering
    \caption{Comparison with LR and LCR detectors on nuScenes \emph{val} (\textbf{top}) and \emph{test} (\textbf{bottom}) set. The reported LiRaFusion results are using TransHead~\cite{FUTR3D}. No test time augmentation or model ensemble is used. Results for other models are the reported results in the papers so some entries are missing. All values are percentages.}
    \scalebox{0.9}{
    \begin{threeparttable}
\begin{tabular}{c|ccccc}
\hline
Method           & Sensor & NDS           & mAP           & AP (car)      & AP (Moto)     \\ \hline
RadarNet \cite{radarnet}         & LR     & N/A           & N/A           & 84.3          & 53.7          \\
RVF-Net \cite{radar_voxel_fusion}          & LR     & N/A           & N/A           & 54.18         & N/A           \\
RVF-Net \cite{radar_voxel_fusion}           & LCR    & N/A           & N/A           & 54.86         & N/A           \\
DeepFusion \cite{deepfusion_lcr}       & LCR    & N/A           & N/A           & 83.5          & N/A           \\
LiRaFusion (ours)       & LR     & 66.7 & 60.1 & 85.6 & 68.1 \\ \hline
Sparse-PointNet \cite{sparse_pointnet}  & LCR    & N/A           & 48.9          & 71            & 36            \\
Frustum PointNet \cite{high_dimen_frustum_pointnet_lcr} & LCR    & N/A           & 36.6          & 48            & 41            \\
LiRaFusion (ours)       & LR     & 66.2 & 59.5 & 84.7 & 63.3 \\
\end{tabular}
\end{threeparttable}}
\label{tab:comparison_other}
\vspace{-4mm}
\end{table}

\begin{table}[t!]
\caption{Runtime and True Positive (TP) metrics for experiments evaluated with TransHead~\cite{FUTR3D}. Runtime is measured using frames per second (FPS). Lower is better across all metrics.}
\scalebox{0.88}{
\begin{threeparttable}
\begin{tabular}{c|c|ccccc}
\hline
Method          & Runtime & mATE           & mASE           & mAOE           & mAVE           & mAAE           \\ \hline
FUTR3D-LO       & 6.0 FPS     & \textbf{0.342} & \textbf{0.265} & 0.321          & 0.276          & 0.193          \\
FUTR3D-LR       & 5.6 FPS    & 0.347          & 0.267          & 0.308          & 0.256          & 0.189          \\
EZFusion-LR$^*$ & 5.9 FPS    & \textbf{0.342} & 0.266          & 0.315          & 0.267          & 0.196          \\
LiRaFusion (ours) & 5.5 FPS    & 0.346          & 0.267          & \textbf{0.298} & \textbf{0.240} & \textbf{0.186}
\end{tabular}
\begin{tablenotes}
\footnotesize
\item Abbreviations: mean Average Translation Error (mATE), mean Average Scale Error (mASE), mean Average Orientation Error (mAOE), mean Average Velocity Error (mAVE), mean Average Attribute Error (mAAE).
\end{tablenotes}
\end{threeparttable}}
\label{tab:TP_futr3d}
\vspace{-7mm}
\end{table}

In addition to the comparison with the most recent state-of-the-art LR detectors \cite{FUTR3D, ezfusion} shown above, we compare LiRaFusion (with TransHead) with other LR or LCR detectors for a complete overview. Table \ref{tab:comparison_other} demonstrates that LiRaFusion has consistent improvement over existing LR and LCR detectors. It is worth mentioning that many detectors in Table \ref{tab:comparison_other} have to enforce a partial class setting, or use the extra camera modality, while LiRaFusion trains on the complete class setting with the LR sensor configuration and outperforms all the LR and LCR detectors.

We additionally report the model runtime and the True Positive (TP) metrics defined in \cite{nuscenes} evaluated with the complete class setting in Table \ref{tab:TP_futr3d}. The table shows LiRaFusion is comparable with other baselines in runtime, which is measured on the same desktop with an RTX A6000 GPU. We can also see LiRaFusion is comparable with the best method in terms of translation and scale error, while achieving the lowest error in estimating orientation, velocity and attribute. We argue the significant reduction in orientation and velocity error comes from the effective fusion of the Doppler information from radars.

\begin{table}[b]
\caption{Performance by object distance of experiments with TransHead~\cite{FUTR3D}. We report the mAP separately in $3$ ranges: $0m-20m$, $20m-30m$ and $30m-50m$. With increasing object distance, LiRaFusion demonstrates higher gain over FUTR3D-LO as radars complement LiDAR at far locations where LiDAR suffers from data sparsity. All values are percentages.}
\centering
\begin{threeparttable}
\begin{tabular}{c|ccc}
\hline
                         & \multicolumn{3}{c}{mAP}                                                \\
\multirow{-2}{*}{Method} & $[0m,20m]$                                  & $[20m,30m]$ & $[30m,50m]$ \\ \hline
FUTR3D-LO                & 73.86                                       & 55.2        & 29.93      \\
LiRaFusion (ours)          & 74.14 {\color[HTML]{FD6864}$\uparrow 0.28$} & 55.88 {\color[HTML]{FD6864}$\uparrow 0.68$}      & 31.73 {\color[HTML]{FD6864}$\uparrow 1.8$}    
\end{tabular}
\end{threeparttable}
\label{tab:distance_futr3d}

    \centering
    \caption{Performance by weather conditions of experiments with TransHead~\cite{FUTR3D}. We report results of two weather conditions: Sunny and Rainy. The grouping strategy is based on the official scene description entries in nuScenes \cite{nuscenes}. There are in total $968$ rainy frames out of $6019$ frames.
All values are percentages.}
    \scalebox{0.86}{
    \begin{threeparttable}
\begin{tabular}{c|cc|cc}
\hline
& \multicolumn{2}{c|}{NDS} & \multicolumn{2}{c}{mAP}         \\
\multirow{-2}{*}{Method} & Sunny & Rainy & Sunny & Rainy \\ \hline

FUTR3D-LO & 65.55 & 65.39 & 59.28 & 57.32 \\
LiRaFusion (ours) & 66.42 {\color[HTML]{FD6864}$\uparrow 0.87$} & 67.15 {\color[HTML]{FD6864}$\uparrow 1.76$} & 59.93 {\color[HTML]{FD6864}$\uparrow 0.65$} & 59.35 {\color[HTML]{FD6864}$\uparrow 2.03$}
\end{tabular}
\end{threeparttable}}
\label{tab:weather_conditions}
\end{table}

We also group predictions based on object distances from the ego-vehicle and weather conditions to further break down the performance boost, as this directly represents the improvement from radars. Table \ref{tab:distance_futr3d} shows the performance breakdown based on the distance. We find the performance gain increases with increasing distance. This finding matches the expectation that radars complement LiDAR by providing additional information at distant locations where the LiDAR returns become sparser. Table \ref{tab:weather_conditions} shows that LiRaFusion achieves notable improvement over the LO baselines in rainy scenes, where LiDAR is generally believed to have reduced detection capability \cite{seeing_through_fogs, radarnet}. This finding validates the importance of fusing radars and leveraging their robustness across different weather conditions.

\subsection{Qualitative Results}

\begin{figure}[h]
\begin{center}
  \includegraphics[width=1.0\linewidth]{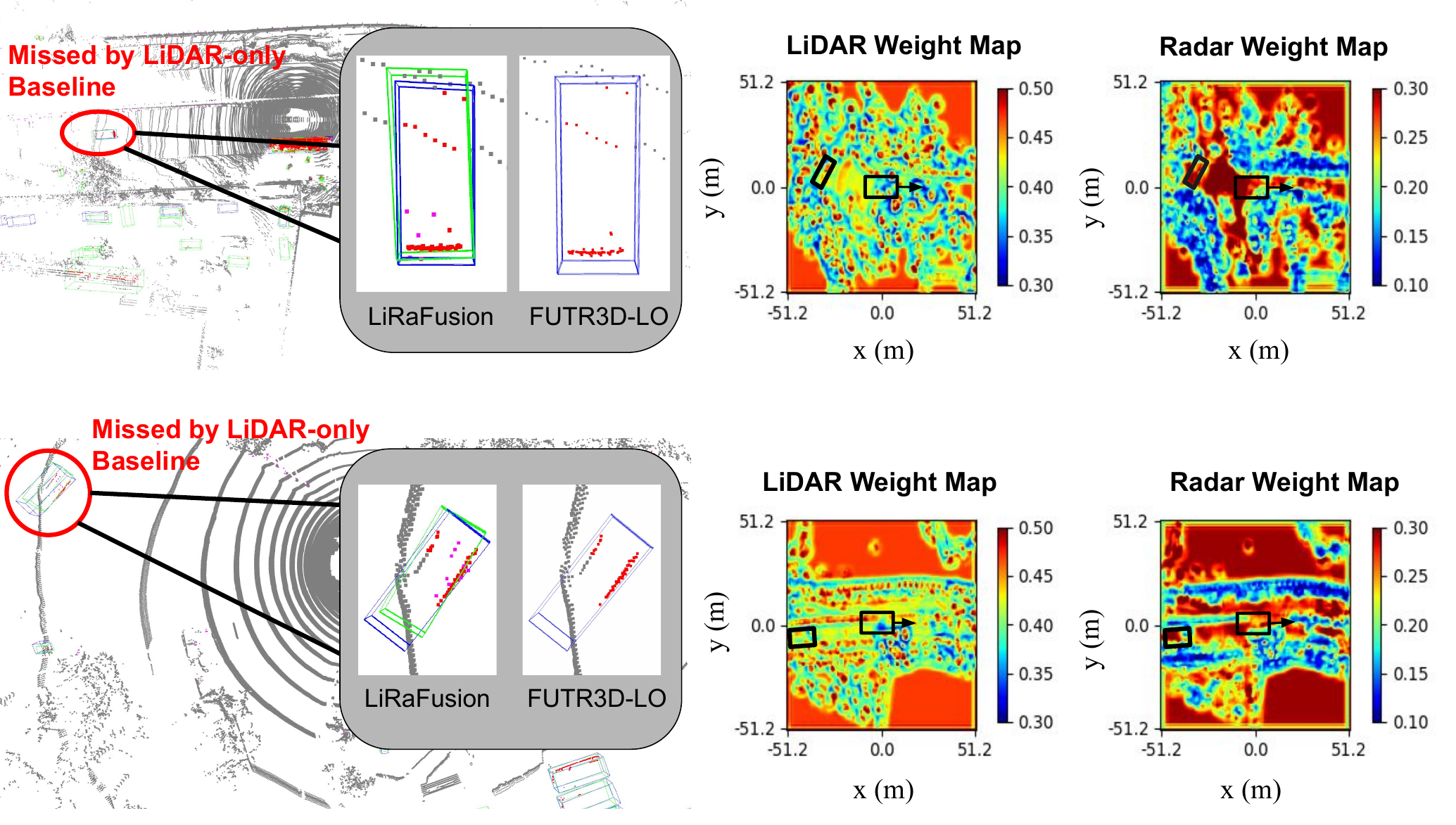}
  \caption{Example bounding box predictions and corresponding weight maps. We present two frames in which LiRaFusion correctly detects a car (highlighted with a red circle) that is missed by the baseline LO detector. We also show a zoomed-in view in which we label radar points in {\color{magenta}magenta}, and LiDAR points in {\color{gray}gray} or {\color{red}red} (if they reside in a bounding box). We show ground truth bounding boxes in {\color{blue}blue} and predictions in {\color{green}green}. In the visualization of weight maps, the {\color{black}black} bounding box with arrow denotes the ego-vehicle. Boxes without an arrow denote the highlighted missed car object. Best viewed in color and zoomed-in.
}
  \label{fig:qualitative_preds}
\end{center}
\vspace{-8mm}
\end{figure}
Figure~\ref{fig:qualitative_preds} shows qualitative comparison of LiRaFusion and FUTR3D-LO by presenting the predictions along with the ground truth bounding boxes. These results show that the radar sensor contributes several measurement points (shown in magenta) for a car that was previously missed by FUTR3D-LO, which only uses LiDAR data. The radar and LiDAR points are used effectively by LiRaFusion and the prediction by LiRaFusion aligns well with the ground truth.

We also show the corresponding adaptive weights on LiDAR and radar feature maps in Fig.~\ref{fig:qualitative_preds}.
We see that the learned weight on the LiDAR feature map is generally higher than that in the radar feature map, which meets the expectation that LiDAR is the preferred sensor for object detection in terms of the density and geometric accuracy. We also notice that in the radar weight map, locations with larger distance to the ego-vehicle have relatively larger weight. This finding matches our intuition that incorporating radar data provides more information for long-range detection capabilities \cite{radarnet, radar_handbook, survey_deep_radar}. Basically, the smaller radar weight at the near locations means that the detector learns to be more dependent on LiDAR at places where it has dense returns, while trusting radar returns at farther locations where LiDAR data has reduced density. Note that the radar weight within the black bounding box in Fig.~\ref{fig:qualitative_preds} is relatively large, which means that the proposed gated network is capable of learning to use the expert feature maps adaptively to let the LiDAR and the radar complement each other in a more effective way.

\begin{table}[h!]
\caption{Ablation study of fusion module.}
    \begin{threeparttable}
\begin{tabular}{c|cccc}
\hline
Method            & NDS $\uparrow$   & mAP $\uparrow$   & mAVE $\downarrow$  & AP (car) $\uparrow$ \\ \hline
FUTR3D-LO         & 65.74 & 59.39 & 0.276 & 84.3     \\
LiRaFusion-early  & 65.95 & 59.11 & 0.263 & 85.4     \\
LiRaFusion-middle & 66.59 & 59.45 & 0.241 & 85.3     \\
LiRaFusion        & \textbf{66.69} & \textbf{60.11} & \textbf{0.240} & \textbf{85.6}    
\end{tabular}
\end{threeparttable}
\label{tab:ablation_early_middle}
\vspace{-6mm}
\end{table}
\subsection{Ablation Studies}

We study the contribution of each fusion module to the overall performance. We denote LiRaFusion-early as the model with the early fusion module only and LiRaFusion-middle as the model with the middle fusion module only. We report the results of these models in addition to the baseline FUTR3D-LO. As shown in Table~\ref{tab:ablation_early_middle}, both LiRaFusion-early and LiRaFusion-middle achieve improvement over the baseline in most metrics. When the early fusion and middle fusion modules are combined together, the performance of the combined model is further enhanced. 

When designing the adaptive gated network used in the middle fusion module, we improve over the existing network design in \cite{3dcvf, GIF} that has a constant weight (referred as channel-constant) for all features at one location in the bird's-eye-view (BEV) feature map. Since the $z$ dimension and original feature dimension are merged together to form the BEV feature map, a specific weight for each feature dimension could help to utilize the spatial knowledge. Inspired by this, we propose a channel-specific weight map in the gated network. As shown in Table \ref{tab:ablation_gated_network_design}, the proposed channel-specific gated network outperforms the original network design, which validates the effectiveness of our improvement over the original gated network in \cite{3dcvf, GIF}.

\begin{table}[]
    \centering
    \caption{Ablation study of gated network design.}
    \scalebox{1.0}{
\begin{threeparttable}
\begin{tabular}{c|cc}
\hline
Method            & NDS $\uparrow$   & mAP $\uparrow$ \\ \hline
LiRaFusion-middle (Channel-constant)         & 66.01 & 58.84     \\
LiRaFusion-middle (Channel-specific)        & \textbf{66.59} & \textbf{59.45}    
\end{tabular}
\end{threeparttable}}
\label{tab:ablation_gated_network_design}
\vspace{-6mm}
\end{table}

\subsection{LiDAR-Camera-Radar Fusion}
We explore the potential of LiRaFusion to fuse LiDAR, camera, and radar for 3D object detection as they are the common sensing modalities on modern AVs. Since most object detectors follow the backbone-neck-head paradigm, LiRaFusion can be applied to many LiDAR-camera (LC) detectors by replacing the LiDAR backbone to enable LCR fusion. FUTR3D \cite{FUTR3D} supports LC configuration (FUTR3D-LC) and is one of the state-of-the-art LC detectors. By replacing its LiDAR backbone to LiRaFusion, we implemented the LCR model referred as LiRaFusion-LCR. We directly compare LiRaFusion with FUTR3D-LC to evaluate the scalability of LiRaFusion. Table \ref{tab:LCR_fusion} shows that LiRaFusion-LCR achieves further improvement over FUTR3D-LC. The results also demonstrate that radars can complement the LC configuration, which reinforces the importance of fusing radar data in modern object detectors. As the main focus of this project is on LR fusion, we leave more experiments on LCR fusion for future work, and hope our work can inspire more research on fusing radars with other sensors to improve perceptual capabilities of AVs.

\begin{table}[h]
    \centering
    \caption{Results with LCR fusion and TransHead~\cite{FUTR3D}.}
\scalebox{0.74}{
\begin{threeparttable}
\begin{tabular}{c|ccccccc}
\hline
Method & NDS $\uparrow$ & mAP $\uparrow$ & mATE$\downarrow$ & mASE$\downarrow$ & mAOE$\downarrow$ & mAVE$\downarrow$ & mAAE$\downarrow$ \\ \hline
FUTR3D-LC & 68.0 & 64.2 & 0.350 & \textbf{0.259} & 0.304 & 0.305 & 0.193 \\
LiRaFusion-LR & 66.69 & 60.11 & 0.346 & 0.267 & \textbf{0.298} & \textbf{0.240} & 0.186 \\
LiRaFusion-LCR & \textbf{68.65} & \textbf{64.76} & \textbf{0.345} & \textbf{0.259} & 0.309 & 0.276 & \textbf{0.181}
\end{tabular}
\begin{tablenotes}
\footnotesize
\item Abbreviations: mean Average Translation Error (mATE), mean Average Scale Error (mASE), mean Average Orientation Error (mAOE), mean Average Velocity Error (mAVE), mean Average Attribute Error (mAAE).
\end{tablenotes}
\end{threeparttable}}
\label{tab:LCR_fusion}
\vspace{-6mm}
\end{table}

\section{Conclusion}
We have proposed a novel LiDAR-radar fusion network, LiRaFusion, to facilitate cross-modality feature extraction for 3D object detection. 
We design a joint voxel feature encoder to extract voxel feature encoding in an early stage. 
We propose an adaptive gated network to further fuse the feature maps from LiDAR and radar by learning modality-adaptive weight maps. Experimental results show that LiRaFusion achieves consistent improvement over existing LiDAR-radar detectors on the nuScenes benchmark. Future work includes applying LiRaFusion to existing LiDAR-camera detectors to further improve over existing LCR detectors, and also extending LiRaFusion to other scene understanding tasks. 

\bibliographystyle{ieeetr}
\bibliography{ref}

\end{document}